\documentclass[letterpaper,10pt,conference]{ieeeconf}  

\IEEEoverridecommandlockouts

\usepackage{xparse}

\usepackage{amsmath} 
\usepackage{amssymb}
\usepackage{amsthm}
\usepackage{amsfonts} 
\usepackage{color}

\usepackage{graphicx}
\graphicspath{{figs/}}
\usepackage{calc}

\IEEEtriggeratref{17}

\usepackage{hyperref}

\usepackage{cleveref}
\crefformat{equation}{(#2#1#3)}
\Crefname{figure}{Fig.}{Figs.}
\crefname{figure}{Fig.}{Figs.}
\crefformat{figure}{#2Fig.~#1#3}
\Crefformat{figure}{#2Fig.~#1#3}
\crefformat{table}{#2Table~#1#3}
\Crefformat{table}{#2Table~#1#3}
\crefformat{section}{#2Sec.~#1#3}
\Crefformat{section}{#2Sec.~#1#3}
\crefformat{problem}{#2Problem~#1#3}
\Crefformat{problem}{#2Problem~#1#3}
\crefname{appsec}{Appendix}{Appendices}

\newtheorem{problem}{Problem}

\usepackage{url}
\usepackage{graphicx}

\usepackage{array}
\usepackage{longtable}
\usepackage{multirow}
\usepackage{float}
\usepackage{caption}
\usepackage{mathrsfs}
\usepackage{listings}
\usepackage{rotating}
\usepackage{overpic}

\usepackage{booktabs}

\usepackage{xparse}

\NewDocumentCommand\bbm{}{ \begin{bmatrix} }
\NewDocumentCommand\ebm{}{ \end{bmatrix} }
\NewDocumentCommand\Vector{m}{ \boldsymbol{\mathbf{#1}} }
\NewDocumentCommand\Matrix{m}{ \boldsymbol{\mathbf{#1}} }

\NewDocumentCommand\Norm{m}{\left\Vert#1\right\Vert }

\NewDocumentCommand\Transpose{}{\mathsf{T}}

\NewDocumentCommand\Real{}{ \mathbb{R} }

\NewDocumentCommand\ZeroMatrix{}{ \Matrix{0} }
\NewDocumentCommand\IdentityMatrix{}{ \Matrix{I} }

\NewDocumentCommand\CoordinateFrame{m}{ \underrightarrow{\Matrix{\mathcal{F}}}_{#1} }

\NewDocumentCommand\NumMeasurements{}{ K }

\NewDocumentCommand\Skew{m}{[#1]^\times}
\NewDocumentCommand\SPDMatrices{}{ \mathcal{S}_{++} }
\NewDocumentCommand\Rank{m}{ \mathrm{rank}(#1) }
\NewDocumentCommand\Ones{}{ \Vector{1} }

\NewDocumentCommand\Mass{}{ m }
\NewDocumentCommand\CenterOfMass{}{ \Vector{c} }
\NewDocumentCommand\InertiaMatrix{}{ \Matrix{J} }

\NewDocumentCommand\Gravity{}{ g }
\NewDocumentCommand\Force{}{ f }
\NewDocumentCommand\Torque{}{ \tau }
\NewDocumentCommand\Position{}{ p }
\NewDocumentCommand\InertialParameters{}{ \Vector{\phi} }
\NewDocumentCommand\Acceleration{}{ \Vector{a} }
\NewDocumentCommand\Velocity{}{ \Vector{v} }
\NewDocumentCommand\AngularAcceleration{}{ \Vector{\alpha} }
\NewDocumentCommand\AngularVelocity{}{ \Vector{\omega} }
\NewDocumentCommand\Dynamism{}{ \nu }

\NewDocumentCommand\ObjectFrame{}{ \CoordinateFrame{b} }
\NewDocumentCommand\WorldFrame{}{ \CoordinateFrame{w} }
\NewDocumentCommand\SensorFrame{}{ \CoordinateFrame{s} }

\NewDocumentCommand\NewtonEulerMatrix{}{ \Matrix{A}_{\InertialParameters} }
\NewDocumentCommand\DataMatrix{}{ \Matrix{A} }
\NewDocumentCommand\RedModel{}{ \Matrix{\widetilde{A}} }
\NewDocumentCommand\FullModel{}{ \Matrix{A} }

\NewDocumentCommand\avgnorm{+m}{ \overline{\vert\vert\Vector{#1}\vert\vert} }
\NewDocumentCommand\norm{+m}{ \vert\vert\Vector{#1}\vert\vert_2 }
\NewDocumentCommand\TextCOM{}{COM }
\NewDocumentCommand\TextCOMMA{}{COM}

\title{\LARGE \bf Fast Object Inertial Parameter\\ Identification for Collaborative Robots}

\author{Philippe Nadeau$^\ast$, Matthew Giamou$^\dagger$, and Jonathan Kelly$^\ddagger$
\thanks{All authors are with the STARS Laboratory at the University of Toronto Institute for Aerospace Studies, Toronto, Ontario, Canada. {\tt\footnotesize <firstname>.<lastname>@robotics.utias.utoronto.ca}}
\thanks{$^\ast$Philippe Nadeau was supported in part by the Vector Institute Scholarship in Artificial Intelligence.} 
\thanks{$^\dagger$Matthew Giamou is a Vector Institute Postgraduate Affiliate.}
\thanks{$^\ddagger$Jonathan Kelly is a Vector Institute Faculty Affiliate. This research was supported in part by the Canada Research Chairs program.}}

\begin{document} 
	
\maketitle 
\thispagestyle{empty}
\pagestyle{empty}

\begin{abstract}
	Collaborative robots (cobots) are machines designed to work safely alongside people in human-centric environments.
	Providing cobots with the ability to quickly infer the inertial parameters of manipulated objects will improve their flexibility and enable greater usage in manufacturing and other areas.
	To ensure safety, cobots are subject to kinematic limits that result in low signal-to-noise ratios (SNR) for velocity, acceleration, and force-torque data. 
	This renders existing inertial parameter identification algorithms prohibitively slow and inaccurate.
	Motivated by the desire for faster model acquisition, we investigate the use of an approximation of rigid body dynamics to improve the SNR.
	Additionally, we introduce a mass discretization method that can make use of shape information to quickly identify plausible inertial parameters for a manipulated object.
	We present extensive simulation studies and real-world experiments demonstrating that our approach complements existing inertial parameter identification methods by specifically targeting the typical cobot operating regime.
\end{abstract}

\section{Introduction}

The mass, centre of mass (\TextCOMMA), and moments of inertia of an object constitute its inertial parameters and determine how the object behaves when it is manipulated \cite{mason_dynamic_1993}.
Humans are able to predict where an object's centre of mass (COM) is located \cite{lukos_choice_2007} and reason about the object's mass \cite{hamrick_inferring_2016}.
Humans also leverage a haptic perceptual capability called \textit{dynamic touch} to instinctively infer the moments of inertia of a manipulated object \cite{pagano_eigenvectors_1992}.
In order for collaborative robots (cobots) to work safely alongside human workers, the machines should be given a similar perceptual capacity.

\begin{figure}[t]
\centering
\setlength{\fboxsep}{0pt}%
\setlength{\fboxrule}{0.75pt}%
\fbox{\includegraphics[width=1\linewidth-1.5pt]{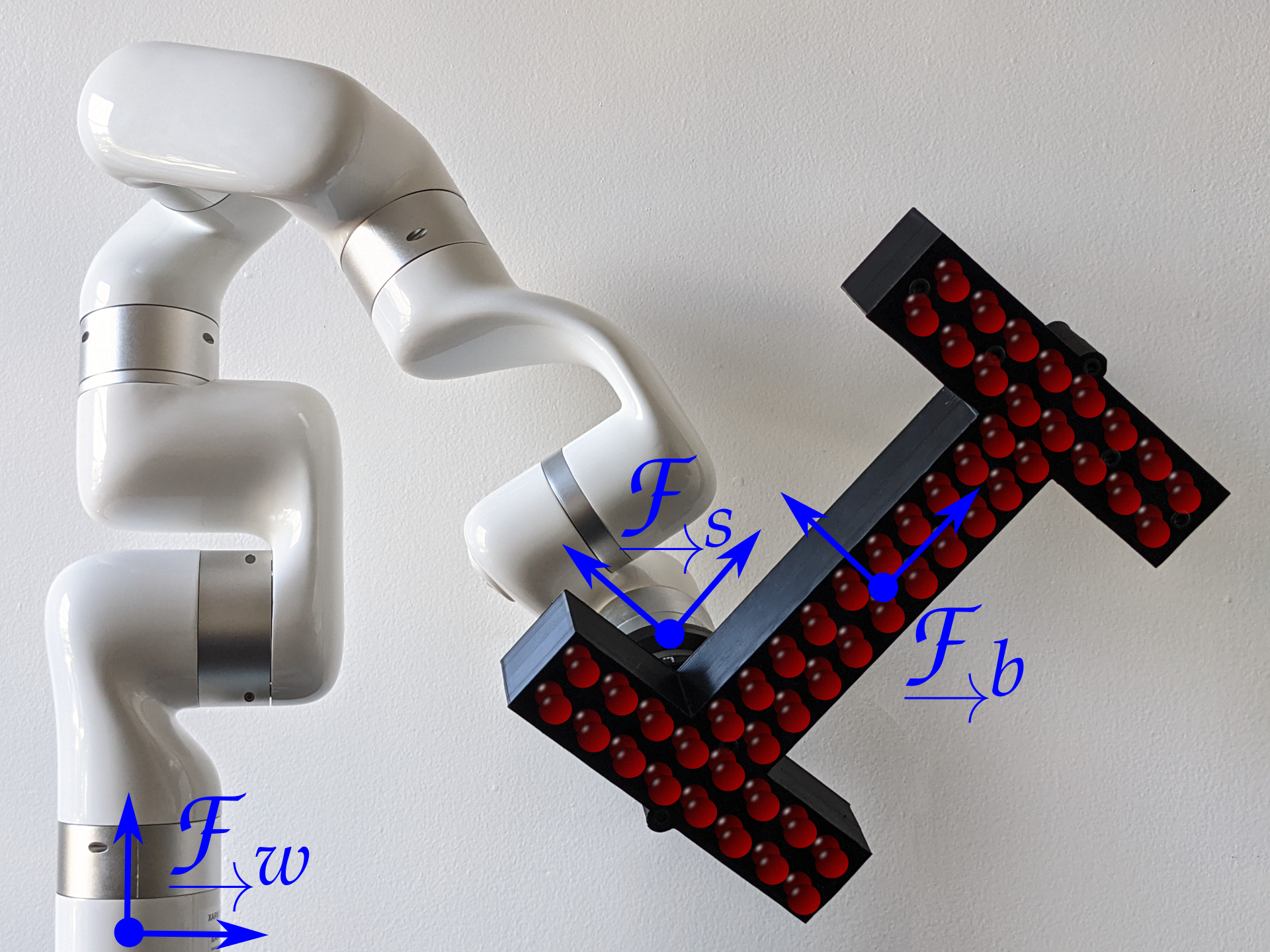}}
\caption{The shape of an object (black) is discretized and point masses are assigned at positions (red) that are fixed relative to the body frame $\ObjectFrame$. The object moves in the inertial frame $\WorldFrame$, while a sensor in frame $\SensorFrame$ measures forces.}
\label{fig:gladyswieldingobject}
\vspace{-5mm}
\end{figure}

In this paper, we study the problem of inertial parameter identification in the context of object manipulation by collaborative robots.
Our first contribution is an analysis of the dynamics of safe cobot operation, which typically hampers identification algorithms due to the low signal-to-noise ratios (SNR) of the velocity, acceleration, and force-torque data.
We investigate an approximation of rigid body dynamics aimed at increasing the SNR and improving the accuracy and convergence rate of parameter identification.
This approximation can easily be integrated into a variety of existing identification algorithms.
However, by using this approximation alone, we show that the identifiability of the full inertia tensor is lost.

Our second contribution is a method to quickly identify all of the inertial parameters by combining the approximate model of the dynamics with the full model through a weighting scheme.\footnote{We provide code, models of our test objects, a video showcasing our algorithm, and supplementary material at \url{https://papers.starslab.ca/fast-inertial-identification/}.}
We propose to leverage knowledge of the shape of the manipulated object; shape information could be captured through the cobot's vision system or from another source.
We base our weighting scheme on a heuristic measure of the `magnitude' of the cobot's motion.
Our method identifies physically consistent values of the inertial parameters and has only a single scalar hyperparameter that can be tuned on a per-robot basis.

\section{Related Work}
\label{sec:related_work}

Unlike quasi-static or kinematic manipulation tasks \cite{mason_toward_2018}, dynamic manipulation requires explicit consideration of the forces applied to the manipulated object \cite{mason_dynamic_1993}.
In this section, we briefly survey inertial parameter identification methods that are most relevant to collaborative robotic manipulation. 
For a recent review of inertial parameter identification for robotic systems and for manipulated objects, we direct our readers to \cite{golluccio2020robot} and  \cite{mavrakis_estimation_2020}, respectively.
	
The classic work of Atkeson, An, and Hollerbach \cite{atkeson_estimation_1986} uses a least squares approach to determine the inertial parameters of an object manipulated by a robot arm. Unfortunately, the low SNR of the force-torque measurements and acceleration signals lead to poor performance.
Kubus, Kroger, and Wahl \cite{kubus_-line_2008} propose to mitigate the effects of noisy and inaccurate acceleration and velocity data, to some extent, by leverage a recursive total least squares formulation (RTLS) and singular value decomposition (SVD).
Farsoni et al.\ \cite{farsoni_real-time_2018} test the RTLS method on a real robotic manipulation task, comparing its performance on acceleration data from multiple different sources. 
Unfortunately, \cite{farsoni_real-time_2018} reports large estimation errors and a slow convergence time, suggesting that the performance of RTLS is limited in noisy, real-world contexts.

The authors of \cite{sousa2014physical} and \cite{traversaro_identification_2016} make the critical observation that inertial parameter identification has been ill-posed traditionally, and that many algorithms converge toward inertial parameters that are unphysical. 
Traversaro et al.\ \cite{traversaro_identification_2016} build upon \cite{sousa2014physical} by stating \textit{sufficient} conditions for physical consistency and by using manifold optimization to constrain the feasible set to physically-consistent solutions.
In \cite{wensing_linear_2017}, a set of linear matrix inequalities (LMIs) that enforce physical consistency without relying on the evaluation of computationally expensive nonlinear functions is proposed.
Leveraging these LMIs, the work in \cite{lee_geometric_2019} regularizes the problem with weighted distance metrics that approximate the geodesic distance between an initial guess and the solution. 

In contrast to classical methods, the authors of \cite{ayusawa_identification_2010} propose a change of paradigm by segmenting a body into points and estimating the mass of each point instead of directly identifying the inertial parameters. The work in \cite{ayusawa_identification_2010} demonstrates that full physical consistency is guaranteed if the point masses are all located in the convex hull of the body, providing a strong theoretical foundation for the approach. 
While we also apply discretization, our work does not assume prior knowledge of the inertial parameters or accurate knowledge of the dynamics, and instead uses information from visual or other perception sources. %

Song and Boularias \cite{song_probabilistic_2020} similarly discretize an object using fixed-size voxels and perform quasi-static exploratory planar manipulation.
Since disentangling the effects of inertia and friction on quasi-static motions is difficult, a probability distribution of the mass and friction coefficient for each voxel is tracked.
Data from tactile and force-torque sensors is used in \cite{sundaralingam_-hand_2021} to estimate the inertial parameters and friction coefficient of a manipulated object, and compare the performance of \cite{atkeson_estimation_1986} with methods considering physical consistency and geodesic distance from an initial guess.

Machine learning is often useful to build models from information-rich sensor data that are hard to interpret.
Leveraging the GelSight tactile sensor \cite{yuan_gelsight_2017}, Wang et al.\ \cite{wang_swingbot_2020} propose an algorithm that learns a physical feature embedding from tactile data during exploratory motion, allowing it to accurately swing an unknown object to a desired orientation.
A learned model is combined with a physics-based model in \cite{zeng_tossingbot_2019} to achieve impressive performance in a pick-and-throw dynamic manipulation task with arbitrary objects grabbed from a cluttered bin.

Having an inertial model for a manipulated object can guide motion and grasp planning.
Mavrakis et al.\ \cite{mavrakis_analysis_2016} use an inertial model of the manipulated object to augment the dynamics of the robot arm and compute the effort required to move the object along a specified trajectory. 

\section{Background}
\label{sec:problem}

The inertial parameters of a three-dimensional rigid body can be grouped as
\begin{multline}
\label{eqn:Params}
\InertialParameters =
\left[\begin{matrix}
\Mass\!\! & \Mass c_x\!\! & \Mass c_y\!\! & \Mass c_z \end{matrix}\right. \\[-1.25mm]
\left.\begin{matrix} J_{xx}\!\! & J_{xy}\!\! & J_{xz}\!\! & J_{yy}\!\! & J_{yz}\!\! & J_{zz} 
\end{matrix}\right]^\Transpose \in \Real^{10},
\end{multline}
where $\Mass \in \Real_+$ is the mass, $\CenterOfMass = [c_x, c_y, c_z]^\Transpose \in \Real^3$ is the location of the \TextCOMMA, and $\InertiaMatrix \in \SPDMatrices^3$ (i.e., the set of 3$\times$3 symmetric positive definite matrices) is the inertia tensor. For more details on rigid body dynamics, we refer the reader to \cite[Ch. 8]{lynch_modern_2017}.
The moments of inertia of an object cannot be directly measured and must be inferred through the Newton-Euler equations
\begin{equation}
	\label{eqn:NewtonEuler}
	\begin{bmatrix}\Vector{\Force}\\ \Vector{\Torque} \end{bmatrix} = \Mass\begin{bmatrix}
		\IdentityMatrix_{3\times 3} & -\Skew{\CenterOfMass}\\ \Skew{\CenterOfMass} & \InertiaMatrix_s 
	\end{bmatrix}\begin{bmatrix}\Acceleration \\ \AngularAcceleration\end{bmatrix} + 
	\begin{bmatrix}
		\Mass\Skew{\AngularVelocity} \Skew{\AngularVelocity} \CenterOfMass\\ \Skew{\AngularVelocity} \InertiaMatrix_s \AngularVelocity
	\end{bmatrix},
\end{equation}
which relate forces $\Vector{\Force} \in \Real^3$ and torques $\Vector{\Torque} \in \Real^3$ to the mass $m \in \Real_+$, \TextCOM $\CenterOfMass \in \Real^3$, inertia tensor $\InertiaMatrix \in \SPDMatrices^3$, linear acceleration $\Acceleration \in \Real^3$, angular acceleration $\AngularAcceleration \in \Real^3$, and angular velocity $\AngularVelocity \in \Real^3$. 
The skew-symmetric operator $\Skew{\cdot}$ transforms a vector $\Vector{u} \in\Real^3$ into a $\Real^{3\times3}$ matrix such that %
$\Skew{\Vector{u}} \Vector{v} = \Vector{u} \times \Vector{v}$.

The Newton-Euler equations can be expressed in matrix form with $\NewtonEulerMatrix \in \Real^{6\times10}$ built from the accelerations and velocities,
\begin{equation}
	\label{eqn:DataMatrix}
	\begin{bmatrix}\Vector{\Force}\\ \Vector{\Torque}\end{bmatrix} = \NewtonEulerMatrix(\Acceleration,\AngularVelocity, \AngularAcceleration) \InertialParameters.
\end{equation}
By observing $\Vector{\Force}$, $\Vector{\Torque}$, $\Acceleration$, $\AngularVelocity$, and $\AngularAcceleration$, a least squares algorithm can be used to infer values for the vector of inertial parameters $\InertialParameters$, even when noise is present in the measured forces and torques \cite{atkeson_estimation_1986}. 
In practice, the following problems arise \cite{kroger200812d}:
\begin{itemize}
	\item the signals from the force-torque sensors located at the end-effector are very noisy, with an unknown bias that drifts over time; and
	\item the amplitudes of the acceleration and velocity signals are small, resulting in low SNRs.
\end{itemize} 
The root-mean-squared amplitude of a signal S with additive noise N can be used to compute the $\text{SNR} = S^2/N^2$. For the specific case of a signal with additive Gaussian noise, the SNR is computed as the mean-squared signal divided by the variance of the noise.

In the absence of knowledge about the noise characteristics, the amplitude of the signals multiplying the inertial parameters in \cref{eqn:NewtonEuler} can quantify the \emph{dynamism} $\Dynamism$ of the motion---this is heuristic measure of the magnitude of the cumulative SNR:
\begin{equation}
	\label{eqn:Dynamism}
	\Dynamism = \bigg\vert\bigg\vert\left[\frac{\norm{a}}{n_1},\frac{\norm{\alpha}}{n_2},\frac{\norm{\omega}}{n_3}\right]\bigg\vert\bigg\vert_2^2~.
\end{equation}
If prior knowledge about the noise is available, each element of the vector in \cref{eqn:Dynamism} can be divided by the corresponding squared noise amplitude. In this work, we do not assume access to such prior knowledge and set $n_1 = n_2 = 1$ and $n_3 = 0.5$.

An extensive dataset of annotated motions and grasps performed by 13 human subjects during a variety of activities of daily living (ADL) is available \cite{saudabayev_human_2018} and quantifies the average velocities and accelerations  at which objects are usually wielded (see \cref{tab:HumanMotionsStats}).
Similar average velocities (0.16 m/s) can be obtained from the Daily Interactive Manipulation Dataset \cite{huang_dataset_2019}, which focuses on cooking tasks.
The ISO standards 10218 \cite{noauthor_robots_nodate} and 15066 \cite{noauthor_robots_nodate-1} define safety bounds for the operation of collaborative robots, including a maximum linear tool centre-point (TCP) speed of 0.25 m/s.

Velocity and acceleration signals can be very noisy if they are obtained through time-differentiation of robot joint positions. 
As shown in \cite{farsoni_real-time_2018}, equipping the robot end-effector with accelerometers does not necessarily yield better measurements, since these sensors can also be inaccurate, especially if they are not properly calibrated.

\subsection{Gravity as a Dominating Force}
\label{sec:MassApprox}
Performing an analysis of objects from the Yale-CMU-Berkeley dataset \cite{calli_benchmarking_2015}, which categorizes 77 items typically used for ADL, allows us to estimate the average mass $\overline{\Mass}$ of the manipulated objects and the average distance $\avgnorm{r}$ from the \TextCOM to the edge of the object.
Excluding from the analysis any object that is flexible, has a mass less than 10 grams, or is clearly intended to be used with both hands, the remaining 63 objects have an average mass of $\overline{m}=0.257$ kilograms and an $\avgnorm{r}=0.081$ metres.\\

\begin{table}
\centering
\caption{Statistics related to human manipulation during activities of daily living (ADL).}
\label{tab:HumanMotionsStats}
\renewcommand{\arraystretch}{1}
\begin{tabular}{lcccc}
	\toprule
	& Symbol & Mean & Median & Unit \\
	\midrule
	Linear velocity &$\Velocity$ & 0.206 & 0.128 & m/s\\
	Angular velocity &$\AngularVelocity$ & 1.08 & 0.703 & rad/s\\
	Linear acceleration &$\Acceleration$ & 1.45 & 1.20 & m/s$^2$\\
	Angular acceleration &$\AngularAcceleration$ & 11.34 & 8.38 & rad/s$^2$\\
	Object mass &$\Mass$ & 0.257 & 0.110 & kg\\
	Object radius &$r$ & 0.081 & 0.068 & m\\
	Duration & $d$ & 6 & 2.3 & s\\
	\bottomrule
\end{tabular}
\vspace{-3mm}
\end{table}

Given the values in \cref{tab:HumanMotionsStats}, we can compare the effect of gravity to the effect of other forces by computing the ratios of the magnitudes of non-gravitational forces and torques to the magnitudes of the total forces and torques applied during manipulation.
These ratios indicate that gravity dominates over other forces by a factor of about five during ADL performed by human subjects.

In turn, we can approximate the mass of the object by the norm of the measured forces  divided by the gravitational acceleration ($\Mass \approx \frac{||\Vector{\Force}||_2}{||\Vector{\Gravity}||_2}$).
Similarly, we can approximate the measured torques $\Vector{\Torque} \approx -\Mass \Skew{\CenterOfMass} \Vector{\Gravity}$
as fully explained by gravity acting on the mass distribution of the object.  
In Section \ref{sec:pmd}, we leverage these approximations to attenuate the impact of the noise associated with the manipulator velocity and acceleration signals.

\subsection{Discretization of the Object Shape and Mass}

The mass distribution of a rigid body can be approximated through the use of a large number of point masses, each with position $\Vector{\Position}_i$ and mass $\Mass_i$.
By doing so, the integral over the volume $V$ of an object of density $\rho$ used to define the $k$th moment of the mass distribution is approximated by a sum over the $n$ point masses:
\begin{equation}
	\label{eqn:MomentApprox}
	\int_{V} \Norm{\Vector{\Position}}^k \rho(\Vector{\Position})\,dV \approx \sum_{i}^{n} \Norm{\Vector{\Position}_i}^k \Mass_i .
\end{equation}
Like force and torque, mass and inertia are additive, meaning that the inertia of a body can be computed as the sum of the inertia of all its constituent point masses.
By combining the approximations of Section \ref{sec:MassApprox} with \cref{eqn:MomentApprox}, the measured torque can be approximated as
\begin{equation} 
\label{eqn:torque_approximation}
	\Vector{\Torque} \approx -\sum_{i}^{n} \Mass_i \Skew{\Vector{\Position}_i} \Vector{g} ~,
\end{equation}
which we expect to be accurate at slow speeds that are representative of manipulation tasks in ADL.
	
\section{The Point Mass Discretization Problem}
\label{sec:pmd}

In this section, we introduce our point mass discretization (PMD) formulation of inertial parameter estimation.
Consider, as in \cref{fig:gladyswieldingobject}, a rigid body attached to reference frame $\ObjectFrame$ and manipulated by a robot whose end-effector pose is described in the inertial reference frame $\WorldFrame$. 
This body is discretized into $n$ point masses, each with constant position $^b\Vector{\Position}_i \in \Real^3$ relative to $\ObjectFrame$, and constant mass $\Mass_i > 0$.
The position $^b\Vector{\Position}_i$ is determined using a sampling algorithm that ensures every point mass is contained within the shape of the object.
While manipulating the object, the robot uses a force-torque sensor attached to reference frame $\SensorFrame$ to take $\NumMeasurements$ measurements.
Since forces and torques are additive, at a particular time step $j$, the sum of the effects of gravity on each point mass should be equal to the $j$th torque measurement. 
Using the approximation in \cref{eqn:torque_approximation}, the reduced model $\RedModel$ at timestep $j$ is written in $\WorldFrame$ as: 
\begin{equation} \label{eqn:data_matrix}
	^w\Vector{\Torque}_{s_j} = 
	-\bbm 
		\Skew{^w\Vector{\Position}_1}\Vector{\Gravity} \cdots \Skew{^w\Vector{\Position}_n}\Vector{\Gravity} 
	\ebm
	\Vector{\Mass} \triangleq \RedModel_{j} \Vector{\Mass}~,
\end{equation}
where $^w\Vector{\Position}_i = {^w\Matrix{R}_{b_j}} {^b}\Vector{\Position}_i + {^w\Vector{t}_{b_j}}$, $^w\Matrix{R}_{b_j}$ and ${^w\Vector{t}_{b_j}}$ are the rotation and translation of $\ObjectFrame$ relative to $\WorldFrame$ for the $j$th measurement, and $\Vector{\Mass} \in \Real^n_+$ is a vector containing all masses.
The optimization problem underlying PMD is defined as follows:
\begin{problem}[Point Mass Discretization] \label{prob:PMD}
	\begin{align}
		\label{eqn:PMDObjFunc}
		&\min_{\Vector{\Mass} \in\Real^n} \\
		&\quad\vert\vert (\Vector{1}-\Vector{w})^\Transpose(\RedModel \Vector{\Mass} - \Vector{b}) \vert\vert_2 + \vert\vert \Vector{w}^\Transpose(\FullModel \Vector{\Mass} - \Vector{b}) \vert\vert_2 \notag + \lambda \vert\vert \Vector{\Mass}\vert\vert_2 \\
		&\quad\text{\emph{s.t.}} \quad \Mass_i \ge 0 \enspace \forall i \in \{1, \ldots, n\}. \notag
	\end{align}
	where
	$\RedModel$ and $\FullModel$ are in $\Real^{6\NumMeasurements \times n}$ and are formed by stacking $\NumMeasurements$ measured reduced and full model submatrices respectively, which are in $\Real^{6\times n}$.
	Wrench vector $\mathbf{b} \in \Real^{6\NumMeasurements}$ is formed by stacking $\NumMeasurements$ wrench measurements $[^w\Vector{\Torque}_{s_j}, ^w\Vector{\Force}_{s_j}]^\Transpose$,
	$\Vector{w} \in \Real^{6\NumMeasurements}$ is a weight vector formed by stacking $\NumMeasurements$ vectors $\Vector{w}_j$  defined below,
	and $\lambda \in \Real_+$ is a regularization factor encouraging homogeneity and ensuring that a unique solution exists.
\end{problem}

At timestep $k$, the weight vector $\Vector{w}_k$ is formed by stacking six identical weight values $w$ that are computed according to
\begin{equation}
	\label{eqn:WeightSched}
	w = \tanh\left(\frac{3\Dynamism}{c_1}\right)~,
\end{equation}
where $\Dynamism \in \Real_+$ is the dynamism defined in \cref{eqn:Dynamism}. 
The scalar factor $s=3$ multiplying $\Dynamism$ makes it easier to tune the hyperparameter $c_1 \in \Real_+$ and defines the level of dynamism required for the full model to account for $100\cdot\tanh(s)$ percent of the objective function  (i.e., 99.5\% if $s=3$). 
In this work, the regularization factor $\lambda$ was set to $0.1$.

Since the objective function is the norm of an affine expression and the constraints are all linear, \cref{prob:PMD} is a convex optimization problem and can therefore be efficiently solved with general-purpose methods~\cite{boyd_convex_2004}.

\section{Experiments} 
\label{sec:experiments}

In this section, we test PMD in a variety of experiments that highlight the impact of the hyperparameters on its performance.
We compare the PMD algorithm with three benchmarks: the ordinary least-squares algorithm  (OLS), the recursive total least-squares algorithm \cite{kubus_-line_2008} (RTLS), and the algorithm using the entropic distance metric defined in \cite{lee_geometric_2019} (GEO).
Data from simulations are leveraged to compare the performance of identification algorithms under varying levels of synthetic noise and at multiple TCP velocities.
In cases where an identification algorithm requires an initial guess, a homogeneous distribution is assumed.
Finally, these algorithms are compared on data obtained through manipulation trajectories performed by a real robot arm on a variety of test objects.
Our findings demonstrate that PMD outperforms benchmark algorithms in scenarios commonly encountered by collaborative robots.
	
\subsection{Experimental Setup}

\begin{figure}
		\centering
		\includegraphics[height=5cm]{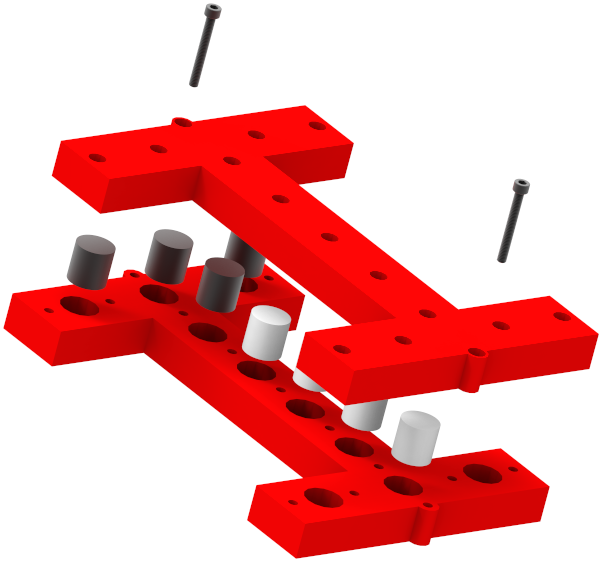}
		\caption{Rendering of the modular test object in the \textit{Hammer} configuration with black steel and white ABS weights.}
		\label{fig:renderingtestobject}
	\vspace{2mm}
	\end{figure}

\begin{figure}
	\centering
	\begin{overpic}[width=1\linewidth]{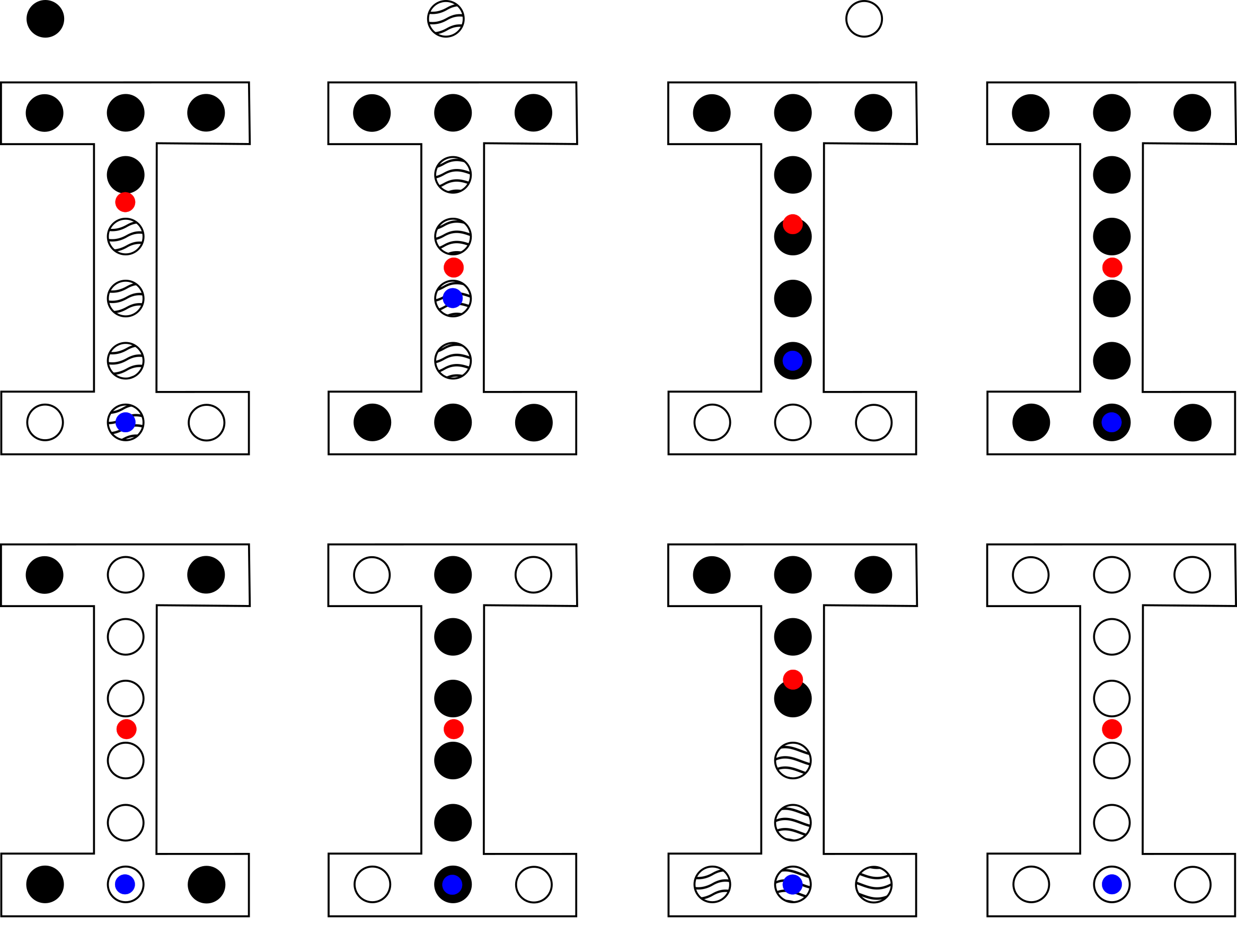}
		\put(6,74){\small Steel Weight}
		\put(38,74){\small ABS Weight}
		\put(72,74){\small No Weight (Air)}
		\put(0.5,36){\small (a) Hammer}
		\put(28.5,36){\small (b) Barbell}
		\put(59,36){\small (c) Tee}
		\put(82,36){\small (d) Uniform}
		\put(1,-1.5){\small (e) Corners}
		\put(30,-1.5){\small (f) Rod}
		\put(52.5,-1.5){\small (g) Half-N-Half}
		\put(83,-1.5){\small (h) Empty}
	\end{overpic}%
	\vspace{3mm}
	\caption{Eight weight configurations used to benchmark the identification algorithms. The \TextCOM (red point) and TCP location (blue point) are shown. The steel and ABS weights are 101 and 17 grams each, respectively.}
	\label{fig:testobjectsconfigs}
\end{figure}
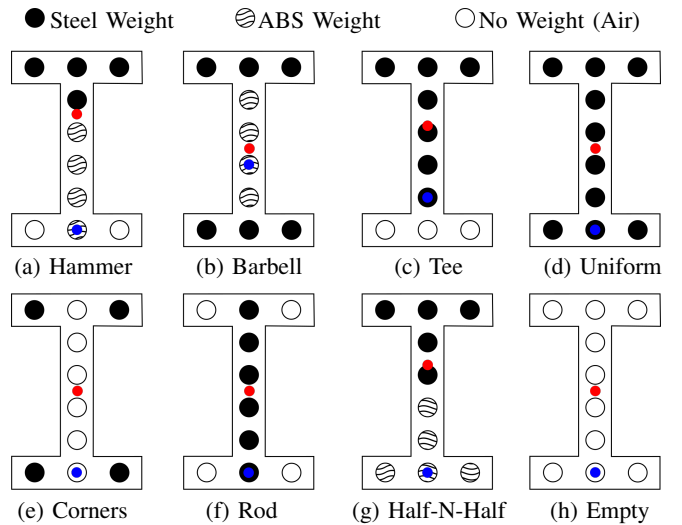

We built a modular test object (shown in \cref{fig:renderingtestobject}) consisting of a 3D-printed PLA structure with holes allowing the placement of weights or bolts. 
Various configurations (shown in \cref{fig:testobjectsconfigs}) can be obtained by choosing the locations of the weights, and precise values of the inertial parameters can easily be computed for any configuration. 
	
\begin{table}[b]
	\caption{Standard deviations of zero-mean Gaussian distributions used to generate noise for simulation experiments.}
	\label{tab:noiseLevels}
	\centering
	\renewcommand{\arraystretch}{1.1}
	\begin{tabular}{c c c c c}
		\toprule
		& Ang. Accel. & Lin. Accel. & Force & Torque\\
		\midrule
		Low Noise 		& 0.25 & 0.025 & 0.05 & 0.0025\\
		Moderate Noise  & 0.5 & 0.05 & 0.1 & 0.005\\
		High Noise 		& 1 & 0.1 & 0.33 & 0.0067\\
		\bottomrule
	\end{tabular}
\end{table}
	
To ensure a fair comparison, algorithms are compared on data obtained with the same motion trajectory.
Inspired by \cite{kubus_-line_2008}, the motion of the end-effector is determined by commanding the three robot joints proximal to the end-effector follow sinusoidal trajectories.
Each trial used the same 35-second motion trajectory obtained by joints following $\alpha_n + 45^{\circ}\sin\left(2\pi f_n t/240\right)$ where $\alpha_n$ is the $n$-th joint's initial position, $f_n$ is $[0,0,0,0.1,0.13,0.16]$ for joints 1 to 6 respectively, and $t$ is the time elapsed in seconds.

\subsection{Error Metrics}

To compare the performance of the algorithms, an adequate error metric needs to be selected.
In the system identification literature, some authors use the standard relative error to put the absolute estimation error in perspective \cite{kubus_-line_2008, farsoni_real-time_2018}.
Although simple and straightforward, this metric breaks when the reference value approaches zero, which occurs frequently with inertial parameters.
Other studies have expressed the error in terms of the distance between the identified inertial parameters and the true parameters by leveraging knowledge of the underlying manifold \cite{traversaro_identification_2016, lee_geometric_2019}.
This unitless distance (and metric) is useful for comparing methods or training learning algorithms, but can be difficult to interpret in practice.

In this work, we propose error metrics (\crefrange{eqn:ErrorMetricMass}{eqn:ErrorMetricInertia}) that are: (i) scale-invariant with respect to the mass and size of the manipulated object, and (ii) work for all physically plausible scenarios.
The length $a^{(i)}$ of the bounding box of the object along axis $i$ is used to take its size into account. 
The Kronecker delta $\delta_{i,j}$ is used to produce a different formula for diagonal ($i=j \rightarrow \delta_{i,j} = 1$) and off-diagonal terms ($i\neq j \rightarrow \delta_{i,j} = 0$) of the inertia tensor. 
The $\hat{x}^{(i)}$ and $\hat{x}^{(i,j)}$ symbols respectively designate the estimation of the $i^{\text{th}}$ element of a vector and of the $(i,j)^{\text{th}}$ element of a matrix. 
\begin{align}
	\label{eqn:ErrorMetricMass}
	{e}_{m} &= \left| \frac{\hat{x} - x}{\Mass} \right| \cdot 100\% \\
	\label{eqn:ErrorMetricCoM}
	{e}_{COM_{i}} &= \left| \frac{\hat{x}^{(i)} - x^{(i)}}{a^{(i)}} \right| \cdot 100\% \\
	\label{eqn:ErrorMetricInertia}
	{e}_{J_{ij}} &= \left| \frac{ (\hat{x}^{(i,j)} - x^{(i,j)}) \cdot 100\%}{\frac{\Mass}{12}\cdot \left( \delta_{i,j}\left( a^{{(i)}^2} + a^{{(j)}^2} + a^{{(k)}^2} \right) - a^{(i)} a^{(j)} \right)} \right|
\end{align}
A scalar average error associated with the \TextCOM can be obtained by averaging the errors along the three axes. Since $\InertiaMatrix$ is symmetric, the scalar average error is computed by averaging the six errors on and above the main diagonal.
	
\subsection{Simulation Experiments}

\begin{figure}
	\centering
	\includegraphics[trim=0cm 0cm 0cm 0cm,clip,width=\linewidth]{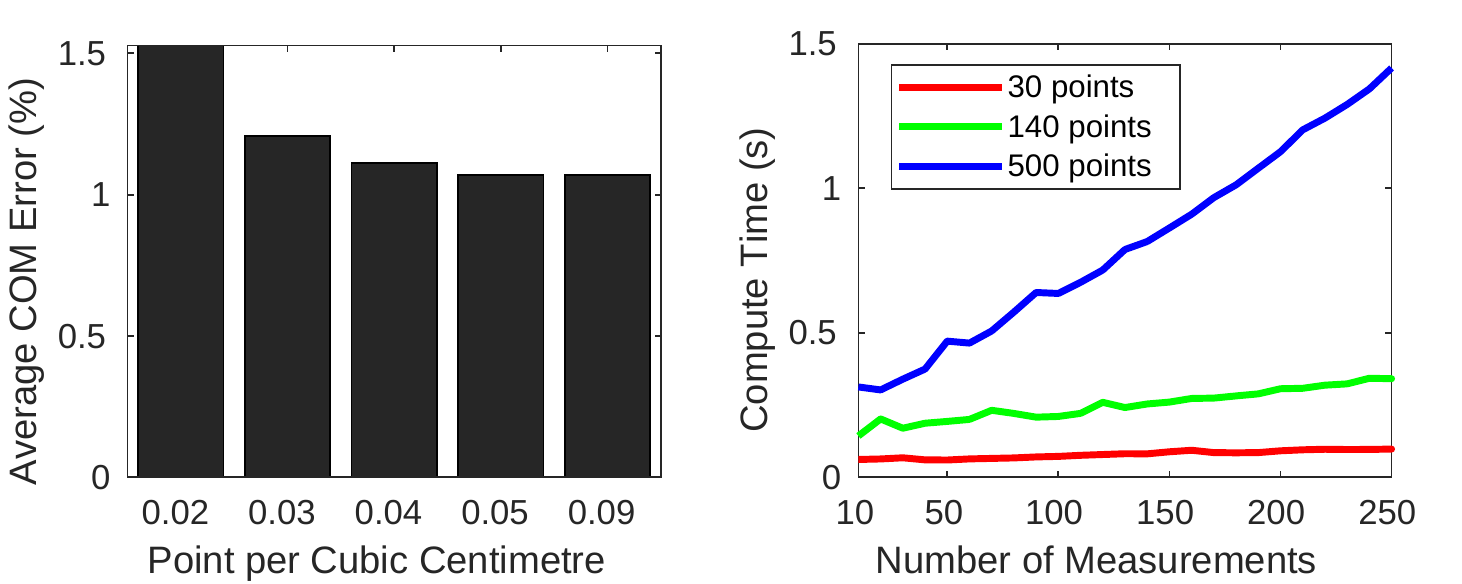}
	\caption{\textit{Left}: Accuracy improves, to some extent, with a greater point mass density. \textit{Right}: Impact of the number of point masses on the compute time as more data is gathered.}
	\label{fig:PointDensity}
\vspace{-4mm}
\end{figure}

\begin{figure}
	\centering
	\includegraphics[trim=1cm 0cm 1cm 0.1cm,clip,width=\linewidth]{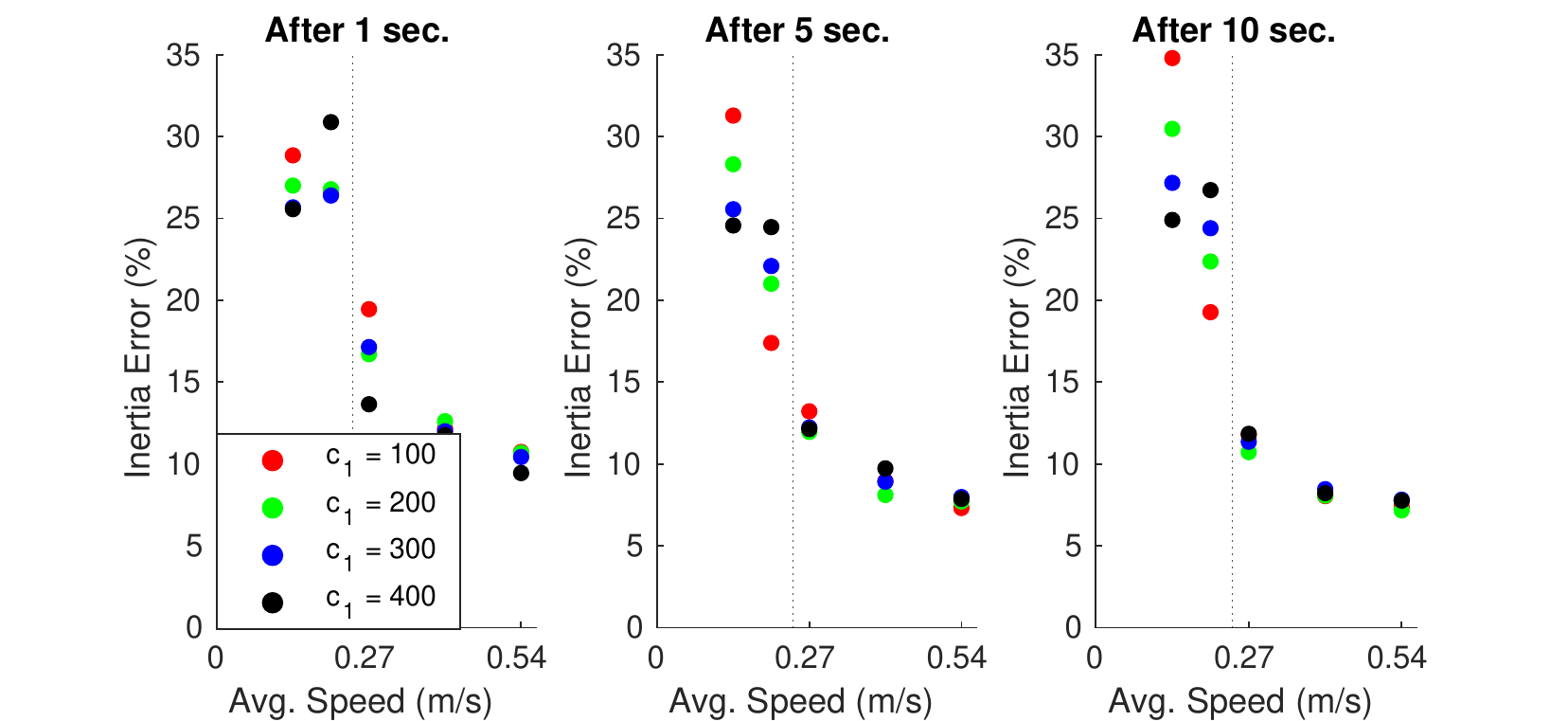}
	\caption{PMD accuracy over time as a function of average TCP speed for different values of the $c_1$ parameter.}
	\label{fig:C1Param}
\vspace{-4mm}
\end{figure}

PMD has a number of hyperparameters that can be tuned to affect accuracy and the compute time required to converge to a solution. 
Simulations were performed for eight different objects (shown in \cref{fig:testobjectsconfigs}) that were voxelized to produce a variety of point mass densities (shown in \cref{fig:PointDensity}), and with a number of different values for $c_1$ in \cref{prob:PMD}.
As a result, 328 simulations 
provide us with an overview of each parameter's impact on identification accuracy and compute time (\cref{fig:PointDensity} and \cref{fig:C1Param}). A point mass density of $0.04$ $\text{points}/\text{cm}^3$ (56 points with our object) and $c_1=300$ were used for these experiments.

\begin{figure}[b]
	\vspace{-4mm}
	\centering
	\includegraphics[trim=1cm 0cm 1cm 0cm,clip,width=\linewidth]{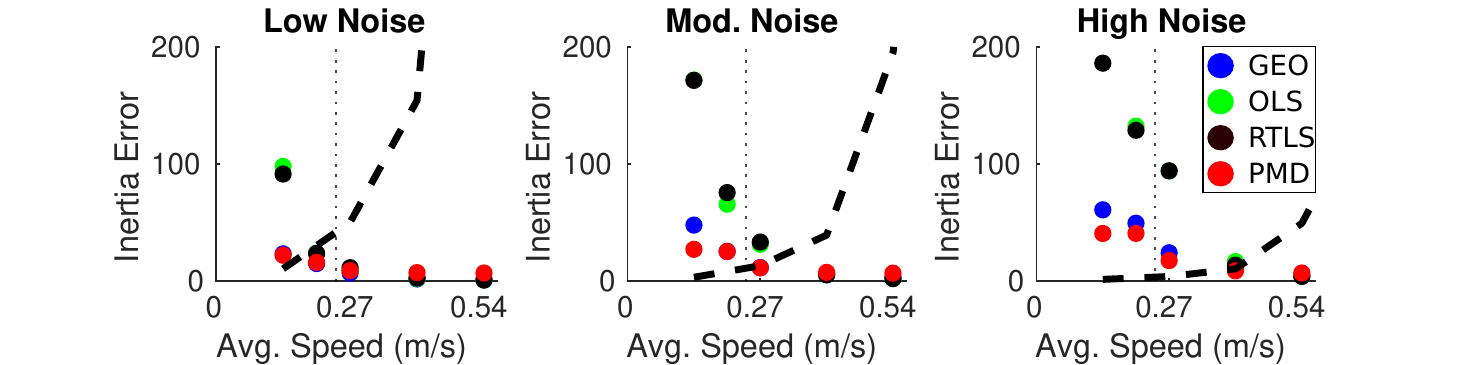}
	\caption{Average performance after 10 sec. for three noise levels and five velocities. The vertical line is ISO 10218-1 max. TCP velocity and the the bold dashed line is the SNR.}
	\label{fig:SimNoise}
\end{figure}

\begin{figure}
	\centering
	\includegraphics[trim=1cm 0cm 1cm 0cm,clip,width=\linewidth]{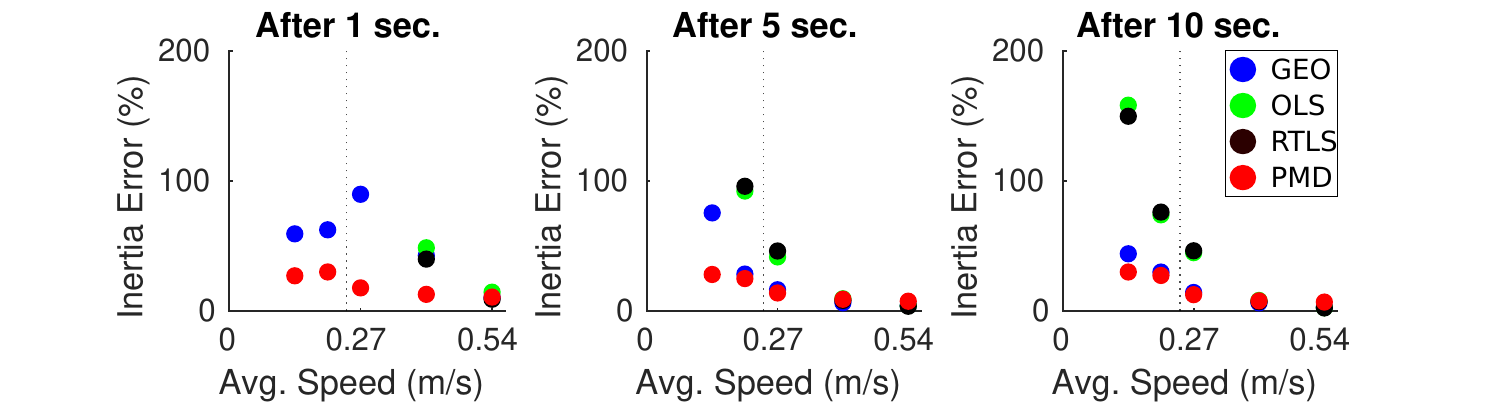}
	\caption{Average performance of each algorithm over time for the moderate noise level and at five end-effector velocities. The dotted line is ISO 10218-1 maximal TCP velocity.}
	\label{fig:SimTime}
\vspace{-4mm}
\end{figure}

To ensure a fair comparison of the identification algorithms, experiments are conducted with the wielded object attaining average angular velocities of 1, 1.5, 2, 3, and 4 rad/s, and average linear velocities of 0.14, 0.2, 0.27, 0.41, and 0.54 m/s.	
For each object in \cref{fig:testobjectsconfigs} and for each velocity, simulated experiments were performed under four different noise levels for a total of 640 experiments.
All algorithms based on optimization problems (i.e., PMD, OLS, GEO) used the MOSEK \cite{andersen2000mosek} solver on the same machine for a fair comparison.
The specifications of the Robotiq FT-300 sensor were used to select the standard deviations of the zero-mean Gaussian noise that was added to the signals as described in \cref{tab:noiseLevels}.
For each experiment, the average error of the inertia tensor as computed with \cref{eqn:ErrorMetricInertia} was used to quantify the identification performance.
\Cref{fig:SimNoise} and \cref{fig:SimTime} summarize results from the experiments where noise was added to the signals.

In a realistic manipulation scenario, only a very short amount of time can be devoted to inertial parameter identification. Indeed, the median duration of manipulation tasks during ADL is 2.3 seconds (see \cref{tab:HumanMotionsStats}).
Assuming a standard force-torque sensing frequency of 100 Hz and taking into account the duration of the optimization procedure, approximately 150 observations can be gathered before the algorithm needs to generate its solution.
The average condition number of the scaled regressor matrix $\DataMatrix$ according to \cite{gautier1992exciting} is 147, 68.4, and 58.0 for velocities of 1.0, 1.5, and 2.0 m/s respectively. These moderately low condition numbers suggest that $\DataMatrix$ is relatively well-conditioned in our experiments.
\Cref{tab:fastInference} shows the average accuracy of each algorithm (with PMD using $c_1=300$) for motion trajectories performed at three velocities with moderate noise added to the forces and torques, and with velocity and acceleration signals estimated through Kalman filtering as described in \cite{farsoni2017compensation}.
In this setting, PMD and GEO always produced physically consistent solutions while OLS and RTLS produced implausible solutions most of the time.
	
\begin{table}[b]
	\centering
	\vspace{-2mm}
	\caption{Inertial parameter accuracy comparison after a limited number of observations have been made (i.e., $\sim$150).}
	\label{tab:fastInference}
	\begin{tabular}{p{0.5cm}p{0.25cm}p{0.25cm}p{0.25cm} p{0.25cm}p{0.25cm}p{0.25cm} ccc}
		\toprule
		& \multicolumn{3}{c}{Mass Error (\%)} & \multicolumn{3}{c}{\TextCOM Error (\%)} & \multicolumn{3}{c}{Inertia Tensor Error (\%)}\\
		rad/s& 1.0 & 1.5 & 2.0 & 1.0 & 1.5 & 2.0 & 1.0 & 1.5 & 2.0 \\
		\midrule
		OLS & 1.36 & 2.88 & 3.22 & 34.4 & 44.0 & 36.1 & $>$500 & $>$500 & $>$500\\
		RTLS& 1.84 & \textbf{1.77} & 2.41 & 97.5  & 37.7 & 38.2 & $>$500 & $>$500 & $>$500\\
		GEO & 1.45 & 2.65 & 4.09 & 33.0 & 33.1 & 32.6 & $>$500 & 252    & 109\\
		PMD & \textbf{1.34} & 2.15 & \textbf{1.91} & \textbf{9.83} & \textbf{15.5} & \textbf{16.4} & \textbf{44.1}   & \textbf{43.5}   & \textbf{43.6}\\
		\bottomrule
	\end{tabular}
\end{table}

	As a sanity check, we tested PMD in the scenario where stop-and-go motions are performed (i.e., zero velocity and acceleration while recording data). In this setting, the errors for the mass and \TextCOM were both lower than 0.1\%.
	
\subsection{Robot Experiments}

In order to verify that the PMD algorithm performs well in real settings, we built a physical version of the modular test object to enable verification against ground truth.
Experiments were carried out using a Robotiq FT-300 force-torque sensor that was attached to the end-effector of a uFactory xArm 7 robotic arm. 
The test object was secured to the flange of the sensor (\cref{fig:gladyswieldingobject}) after weights were placed in one of the eight configurations shown in \cref{fig:testobjectsconfigs}. 
For each configuration, the robot performed an oscillating motion in which the three robot joints proximal to the end-effector followed sinusoidal trajectories (similar to those in the simulation experiments). The average angular velocity of the test object was 1.1 rad/s.

\begin{figure}[b]
	\centering
	\includegraphics[trim=0.2cm 0.2cm 1cm 6.1cm,clip,width=1\linewidth]{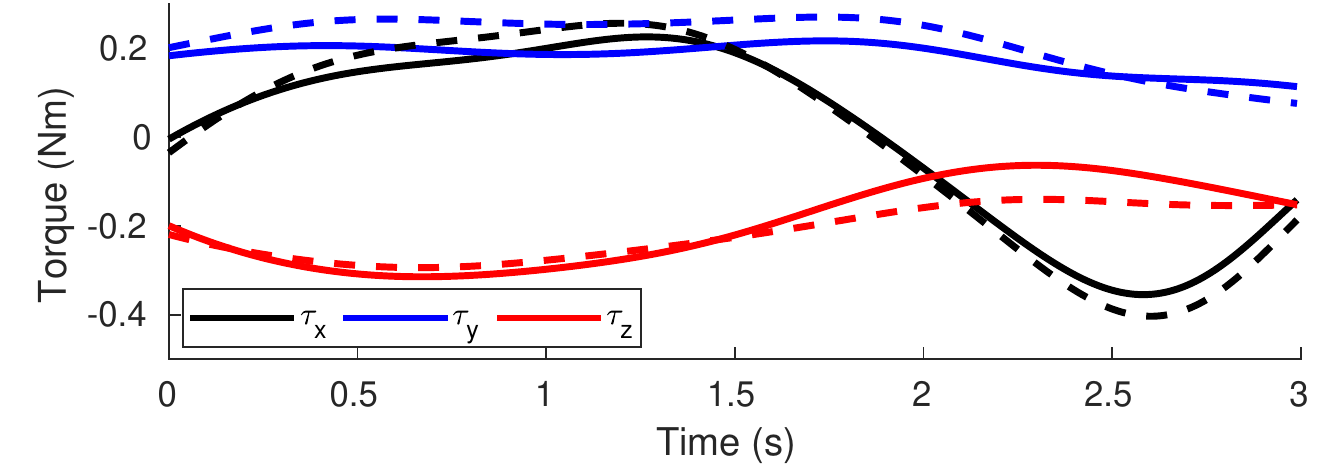}
	\caption{Torques predicted using the true (solid) and estimated (dashed) parameters for the real \textit{Tee} object and for PMD-5K.}
	\label{fig:predictedwrench}
\end{figure}

Akin to the simulation experiments, each identification algorithm was used on data gathered with each object configuration, but the trajectory was limited to 1.5 seconds such that the identification process was faster than the median manipulation duration. PMD was tested with $c_1=5\cdot10^3$ (PMD-5K) and $c_1=25\cdot10^3$ (PMD-25K), and observations were used after the uncertainty estimates from the Kalman filters had stabilized.
For this experiment, the condition number of $\DataMatrix$ is 178---larger than in the simulations but still reasonably small. 
The estimated inertial parameters can be used to predict the effort required to move an object along a given trajectory (see \cref{fig:predictedwrench} for an example with PMD-5K) with reasonable accuracy, enabling effort-based motion planning as done in \cite{mavrakis_analysis_2016}.
The average error across all studied configurations was computed for each algorithm and is shown in \cref{tab:realresults}.
	
\begin{table}[t]
	\centering
	\caption{Average identification errors with the xArm 7 robotic arm and the FT-300 force-torque sensor.}
	\label{tab:realresults}
	\begin{tabular}{p{1.4cm}ccc}
		\toprule
		& Mass (\%) & \TextCOM (\%) & Inertia Tensor (\%)\\
		\midrule
		OLS & \textbf{3.18} & 24.9 & $>$500 \\
		RTLS& 11.7 & $>$500 & $>$500 \\
		GEO & 3.75 & 30.3 & $>$500 \\
		PMD-5K & 4.51 & 8.12  & 36.75\\
		PMD-25K& 4.45 & \textbf{4.50} & \textbf{29.11}\\
		\bottomrule
	\end{tabular}
\vspace{-4mm}
\end{table}

\subsection{Discussion}

As expected, a greater density of point masses leads to a longer compute time. 
Interestingly, the accuracy of the estimate of the COM when using a density of 0.09 points/cm$^3$ was not significantly better than when using a density of 0.04 points/cm$^3$, suggesting that the latter density is a good compromise between accuracy and compute time.
We note that all of the algorithms benchmarked in this work are complementary and that each is suitable in a specific context:
\begin{itemize}
	\item OLS \cite{atkeson_estimation_1986} is the simplest algorithm and performs well with high-speed manoeuvres, but does not guarantee physical consistency;
	\item RTLS \cite{kubus_-line_2008} has the lowest asymptotic error for trajectories with large signal-to-noise ratios, but converges more slowly and does not guarantee physical consistency;
	\item GEO \cite{lee_geometric_2019} performs better than RTLS under high noise and guarantee physical consistency, but requires a good initial guess and a well-tuned regularization factor;
	\item PMD is fast, more robust for small signal-to-noise ratios, and guarantees physical consistency, but requires knowledge of the shape and pose of the object.
\end{itemize}
	
\section{Conclusion}
\label{conclusion}

In this paper, we proposed an inertial parameter estimation method well suited to the velocity regime of cobots during object manipulation.
Our algorithm uses a weighting scheme based on the \emph{dynamism} of the motion; the estimator applies a reduced dynamics model when appropriate to yield a solution quickly.
We rely on a discretization of the object shape, where the requisite information is known or is captured via the cobot's perception system, to guarantee physical consistency of the inertial parameters.
Our proposed algorithm performs well in real-world experiments representative of practical manipulation tasks. %
Promising lines of future work include taking into account uncertainty in the object shape, pose, and inertial parameters in order to guide the cobot's motion during the identification process. 

\bibliographystyle{ieeetr}
\bibliography{MyLibrary}

\begin{thebibliography}{10}

\bibitem{mason_dynamic_1993}
M.~T. Mason and K.~M. Lynch, ``Dynamic manipulation,'' in {\em Proc. 1993
  {IEEE}/{RSJ} {Int.} {Conf.} on {Intelligent} {Robots} and {Systems}}, vol.~1,
  pp.~152--159, 1993.

\bibitem{lukos_choice_2007}
J.~Lukos, C.~Ansuini, and M.~Santello, ``Choice of contact points during
  multidigit grasping: {Effect} of predictability of object center of mass
  location,'' {\em J. Neuroscience}, vol.~27, no.~14, pp.~3894--3903, 2007.

\bibitem{hamrick_inferring_2016}
J.~B. Hamrick, P.~W. Battaglia, T.~L. Griffiths, and J.~B. Tenenbaum,
  ``Inferring mass in complex scenes by mental simulation,'' {\em Cognition},
  vol.~157, pp.~61--76, 2016.

\bibitem{pagano_eigenvectors_1992}
C.~C. Pagano and M.~Turvey, ``Eigenvectors of the inertia tensor and perceiving
  the orientation of a hand-held object by dynamic touch,'' {\em Perception \&
  Psychophysics}, vol.~52, no.~6, pp.~617--624, 1992.

\bibitem{mason_toward_2018}
M.~T. Mason, ``Toward robotic manipulation,'' {\em Annual Review of Control,
  Robotics, and Autonomous Systems}, vol.~1, pp.~1--28, 2018.

\bibitem{golluccio2020robot}
G.~Golluccio, G.~Gillini, A.~Marino, and G.~Antonelli, ``Robot dynamics
  identification: {A} reproducible comparison with experiments on the kinova
  {Jaco2},'' {\em IEEE Robotics \& Automation Magazine}, 2020.

\bibitem{mavrakis_estimation_2020}
N.~Mavrakis and R.~Stolkin, ``Estimation and exploitation of objects’
  inertial parameters in robotic grasping and manipulation: {A} survey,'' {\em
  Robotics and Autonomous Systems}, vol.~124, p.~103374, 2020.

\bibitem{atkeson_estimation_1986}
C.~G. Atkeson, C.~H. An, and J.~M. Hollerbach, ``Estimation of inertial
  parameters of manipulator loads and links,'' {\em The Int. J. Robotics
  Research}, vol.~5, no.~3, pp.~101--119, 1986.

\bibitem{kubus_-line_2008}
D.~Kubus, T.~Kroger, and F.~M. Wahl, ``On-line estimation of inertial
  parameters using a recursive total least-squares approach,'' in {\em Proc.
  2008 {IEEE}/{RSJ} {Int.} {Conf.} on {Intelligent} {Robots} and {Systems}},
  pp.~3845--3852, 2008.

\bibitem{farsoni_real-time_2018}
S.~Farsoni, C.~T. Landi, F.~Ferraguti, C.~Secchi, and M.~Bonfè, ``Real-time
  identification of robot payload using a multirate quaternion-based {Kalman}
  filter and recursive total least-squares,'' in {\em Proc. 2018 {IEEE} {Int.}
  {Conf.} on {Robotics} and {Automation}}, pp.~2103--2109, 2018.

\bibitem{sousa2014physical}
C.~D. Sousa and R.~Cortesao, ``Physical feasibility of robot base inertial
  parameter identification: {A} linear matrix inequality approach,'' {\em The
  Int. J. Robotics Research}, vol.~33, no.~6, pp.~931--944, 2014.

\bibitem{traversaro_identification_2016}
S.~Traversaro, S.~Brossette, A.~Escande, and F.~Nori, ``Identification of fully
  physical consistent inertial parameters using optimization on manifolds,'' in
  {\em Proc. 2016 {IEEE}/{RSJ} {Int.} {Conf.} on {Intelligent} {Robots} and
  {Systems}}, pp.~5446--5451, 2016.

\bibitem{wensing_linear_2017}
P.~M. Wensing, S.~Kim, and J.-J.~E. Slotine, ``Linear matrix inequalities for
  physically consistent inertial parameter identification: {A} statistical
  perspective on the mass distribution,'' {\em IEEE Robotics and Automation
  Letters}, vol.~3, no.~1, pp.~60--67, 2017.

\bibitem{lee_geometric_2019}
T.~Lee, P.~M. Wensing, and F.~C. Park, ``Geometric robot dynamic
  identification: {A} convex programming approach,'' {\em IEEE Trans.
  Robotics}, vol.~36, no.~2, pp.~348--365, 2019.

\bibitem{yuan_gelsight_2017}
W.~Yuan, S.~Dong, and E.~H. Adelson, ``Gelsight: {High}-resolution robot
  tactile sensors for estimating geometry and force,'' {\em Sensors}, vol.~17,
  no.~12, p.~2762, 2017.

\bibitem{wang_swingbot_2020}
C.~Wang, S.~Wang, B.~Romero, F.~Veiga, and E.~Adelson, ``Swingbot: {Learning}
  physical features from in-hand tactile exploration for dynamic swing-up
  manipulation,'' in {\em Proc. 2020 {IEEE}/{RSJ} {Int.} {Conf.} on
  {Intelligent} {Robots} and {Systems}}, (Las Vegas, USA), Oct. 2020.

\bibitem{zeng_tossingbot_2019}
A.~Zeng, S.~Song, J.~Lee, A.~Rodriguez, and T.~Funkhouser, ``Tossingbot:
  {Learning} to throw arbitrary objects with residual physics,'' {\em Proc.
  Robotics: Science and Systems (RSS)}, 2019.

\bibitem{ayusawa_identification_2010}
K.~Ayusawa and Y.~Nakamura, ``Identification of standard inertial parameters
  for large-{DoF} robots considering physical consistency,'' in {\em Proc. 2010
  {IEEE}/{RSJ} {Int.} {Conf.} on {Intelligent} {Robots} and {Systems}},
  pp.~6194--6201, IEEE, 2010.

\bibitem{song_probabilistic_2020}
C.~Song and A.~Boularias, ``A probabilistic model for planar sliding of objects
  with unknown material properties,'' in {\em Proc. 2020 {IEEE}/{RSJ} {Int.}
  {Conf.} on {Intelligent} {Robots} and {Systems}}, Oct. 2020.

\bibitem{sundaralingam_-hand_2021}
B.~Sundaralingam and T.~Hermans, ``In-hand object-dynamics inference using
  tactile fingertips,'' {\em IEEE Trans. Robotics}, vol.~37, no.~4,
  pp.~1115--1126, 2021.

\bibitem{mavrakis_analysis_2016}
N.~Mavrakis, R.~Stolkin, L.~Baronti, M.~Kopicki, M.~Castellani, and {others},
  ``Analysis of the inertia and dynamics of grasped objects, for choosing
  optimal grasps to enable torque-efficient post-grasp manipulations,'' in {\em
  Proc. 2016 {IEEE}/{RAS} 16th {Int.} {Conf.} on {Humanoid} {Robots}},
  pp.~171--178, 2016.

\bibitem{lynch_modern_2017}
K.~M. Lynch and F.~C. Park, {\em Modern {Robotics}: {Mechanics}, {Planning},
  and {Control}}.
\newblock Cambridge University Press, 2017.

\bibitem{kroger200812d}
T.~Kroger, D.~Kubus, and F.~M. Wahl, ``{12D} force and acceleration sensing:
  {A} helpful experience report on sensor characteristics,'' in {\em Proc. 2008
  {IEEE} {Int.} {Conf.} on {Robotics} and {Automation}}, pp.~3455--3462, 2008.

\bibitem{saudabayev_human_2018}
A.~Saudabayev, Z.~Rysbek, R.~Khassenova, and H.~A. Varol, ``Human grasping
  database for activities of daily living with depth, color and kinematic data
  streams,'' {\em Scientific data}, vol.~5, no.~1, pp.~1--13, 2018.

\bibitem{huang_dataset_2019}
Y.~Huang and Y.~Sun, ``A dataset of daily interactive manipulation,'' {\em The
  Int. J. Robotics Research}, vol.~38, no.~8, pp.~879--886, 2019.

\bibitem{noauthor_robots_nodate}
``Robots and robotic devices - {Safety} requirements for industrial robots -
  {Part} 1: {Robots},'' Tech. Rep. ISO 10218-1:2011, ISO, Geneva (2011).

\bibitem{noauthor_robots_nodate-1}
``Robots and robotic devices - {Collaborative} robots,'' Tech. Rep. ISO/TS
  15066:2016, ISO, Geneva (2016).

\bibitem{calli_benchmarking_2015}
B.~Calli, A.~Walsman, A.~Singh, S.~Srinivasa, P.~Abbeel, and A.~M. Dollar,
  ``Benchmarking in manipulation research: {Using} the {Yale}-{CMU}-{Berkeley}
  object and model set,'' {\em IEEE Robotics \& Automation Magazine}, vol.~22,
  no.~3, pp.~36--52, 2015.

\bibitem{boyd_convex_2004}
S.~Boyd and L.~Vandenberghe, {\em Convex optimization}.
\newblock Cambridge University Press, 2004.

\bibitem{andersen2000mosek}
E.~D. Andersen and K.~D. Andersen, ``The mosek interior point optimizer for
  linear programming: {An} implementation of the homogeneous algorithm,'' in
  {\em High performance optimization}, pp.~197--232, 2000.

\bibitem{gautier1992exciting}
M.~Gautier and W.~Khalil, ``Exciting trajectories for the identification of
  base inertial parameters of robots,'' {\em The Int. J. Robotics Research},
  vol.~11, no.~4, pp.~362--375, 1992.

\bibitem{farsoni2017compensation}
S.~Farsoni, C.~T. Landi, F.~Ferraguti, C.~Secchi, and M.~Bonfe, ``Compensation
  of load dynamics for admittance controlled interactive industrial robots
  using a quaternion-based kalman filter,'' {\em IEEE Robotics and Automation
  Letters}, vol.~2, no.~2, pp.~672--679, 2017.

\end{thebibliography}

\appendices
\section{Observability Analysis of the Reduced Model} \label{app:observability_reduced_model}
In this section, we establish that only $m$ and $\CenterOfMass$ are uniquely identifiable in \cref{prob:PMD} when $\Vector{w} = \ZeroMatrix$, which stems from the independence of the approximation in \cref{eqn:torque_approximation} with respect to the inertia tensor $\InertiaMatrix$. 
We begin by proving that $\Rank{\RedModel} \leq 4$. 
We once again assume that the points are not coplanar (i.e., the rank of $\tilde{\Matrix{P}}$ is four). 
Noting the skew-symmetry of the cross product, we write the $j$th measurement's data matrix in \cref{eqn:DataMatrix} as 
\begin{align} \label{eqn:affine_A}
	\RedModel_j &= \Skew{\Vector{\Gravity}} [{^w\Matrix{R}_{b_j}} {^b}\Vector{\Position}_1 + {^w\Vector{t}_{b_j}} \cdots {^w\Matrix{R}_{b_j}} {^b}\Vector{\Position}_n + {^w\Vector{t}_{b_j}}] \notag \\ 
	&= \Skew{\Vector{\Gravity}} {^w\Matrix{R}_{b_j}} [ {^b}\Vector{\Position}_1  \cdots {^b}\Vector{\Position}_n] + \Ones_n^\Transpose \otimes (\Skew{\Vector{\Gravity}}{^w\Vector{t}_{b_j}})\\
	&= \Skew{\Vector{\Gravity}} {^w\Matrix{R}_{b_j}} \Matrix{P} + \Matrix{T}_j, \notag
\end{align}
where $\otimes$ is the Kronecker product.
\Cref{eqn:affine_A} reveals that each $\RedModel_j$ is simply an affine transformation of $\Matrix{P}$.
Therefore, the complete data matrix is simply the affine transformation 
\begin{equation} \label{eqn:rank_of_A}
	\RedModel = \Matrix{G}\Matrix{P} + \Matrix{T},
\end{equation}
where 
\begin{equation}
	\Matrix{G} = \bbm
	\Skew{\Vector{\Gravity}} {^w\Matrix{R}_{b_1}}
	\cdots
	\Skew{\Vector{\Gravity}} {^w\Matrix{R}_{b_\NumMeasurements}}
	\ebm^\Transpose
	\in \Real^{3\NumMeasurements \times 3},
\end{equation}
and
\begin{equation}
	\Matrix{T} = \Ones_n^\Transpose \otimes [(\Skew{\Vector{\Gravity}}{^w\Vector{t}_{b_1}})^\Transpose \cdots (\Skew{\Vector{\Gravity}}{^w\Vector{t}_{b_{\NumMeasurements}}})^\Transpose]^\Transpose \in \Real^{3\NumMeasurements \times n}.
\end{equation}  
Since $\Rank{\Matrix{P}} = 3$, the linear part of \cref{eqn:rank_of_A} has 
\begin{equation} 
	\Rank{\Matrix{G}\Matrix{P}} = \Rank{\Matrix{G}},	
\end{equation}
and $\Rank{\Matrix{T}} = 1$ since its columns are identical.
By the sub-additive property of matrix rank, we have that $\Rank{\RedModel} \leq 4$ as desired. 
Note that for torque measurements taken in a set of non-degenerate poses, the maximum rank of 4 is attained, as $\Rank{\Matrix{G}} = 3$, and the repeated column of $\Matrix{T}$ is not in the range of $\Matrix{G}\Matrix{P}$.\footnote{If we assume that there exist $1 \leq i < j \leq \NumMeasurements$ such that $^w\Matrix{R}_{b_i}$ and $^w\Matrix{R}_{b_j}$ are not rotations about the same axis, then it is easy to see that $\Rank{\Matrix{G}} = 3$. Additionally, if sampled randomly, the repeated column of $\Matrix{T}$ is almost surely not in the range of $\Matrix{G}\Matrix{P}$ and can be selected to ensure that this is not the case.}

Assuming that $\Rank{\RedModel} = 4$, we can show that the kernel of $\RedModel$ does not include changes to $\Mass$ and the \TextCOM $\CenterOfMass$, and that these quantities are therefore identifiable.
Since the range of $\Matrix{G}\Matrix{P}$ and the range of $\Matrix{T}$ only intersect at the origin, 
\begin{align} 
	&\RedModel \delta\Vector{\Mass} = (\Matrix{G}\Matrix{P} + \Matrix{T})\delta\Vector{\Mass} = \Vector{0}\notag \\
	\iff &\Matrix{G}\Matrix{P}\delta\Vector{\Mass} = \Vector{0} \enspace \mathrm{and} \enspace \Matrix{T}\delta\Vector{\Mass} = \Vector{0},
\end{align}
where $\delta\Vector{\Mass} \neq \Vector{0} \in \Real^n$ is a ``perturbation" of the mass vector. 
Since each row of $\Matrix{T}$ is the same value repeated, the sum of the elements of $\delta\Vector{\Mass}$ is zero, and the perturbation $\delta\Vector{\Mass}$ therefore conserves the total mass. 
Similarly, since $\Matrix{G}$ has linearly independent columns, $\Matrix{G}\Matrix{P}\delta\Vector{\Mass} = \Vector{0} \implies \Matrix{P}\delta\Vector{\Mass} = \Vector{0}$, meaning the \TextCOM is invariant to perturbations in the kernel of $\RedModel$:
\begin{equation}
	\sum_{i=1}^n (\Mass_i + \delta\Mass_i) {^b\Vector{\Position}_i} = \Mass{^b\CenterOfMass} + \Matrix{P}\delta\Vector{\Mass} = \Mass{^b\CenterOfMass}.
\end{equation}
Therefore, the ``unobservable subspace" defined by the kernel of data matrix $\RedModel$ contains precisely those changes to the true mass vector $\Vector{\Mass}$ which do not modify the total mass or the weighted \TextCOM $\Mass\CenterOfMass$.

In order to simplify our analysis of the uniqueness of solutions to the reduced model of \cref{prob:PMD}, note that the quadratic program (QP) obtained by squaring the objective function has the same global optima as the unsquared case. 
This QP has a unique solution when the cost is \emph{strictly} convex, or equivalently, its Hessian $\nabla^2 f = \RedModel^\Transpose \RedModel$ is positive definite \cite{boyd_convex_2004}. 
Since, in general, $\Rank{\RedModel} \leq 4 \leq n$, the Hessian $\RedModel^\Transpose \RedModel$ is only positive \emph{semi}definite, and \cref{prob:PMD} does not have a unique solution. 
Adding a regularizing term $\lambda \Norm{\Vector{\Mass}}^2$ to the QP's objective function modifies the Hessian  to $\RedModel^\Transpose \RedModel + \lambda \IdentityMatrix,$ which is positive definite for all $\lambda > 0$.

\end{document}